\documentclass[sigconf,screen,nonacm]{acmart}
\AtBeginDocument{%
  \providecommand\BibTeX{{%
    \normalfont B\kern-0.5em{\scshape i\kern-0.25em b}\kern-0.8em\TeX}}}
\usepackage{booktabs}
\usepackage{epsfig}
\usepackage{enumitem}
\usepackage{multirow}
\usepackage{algorithm,algorithmicx,algpseudocode}
\usepackage{hyperref}
\usepackage{graphicx}
\usepackage{amsmath}
\usepackage{booktabs}
\usepackage{float}
\usepackage{stfloats} 
\usepackage{dsfont}
\usepackage{subcaption}
\usepackage{caption}
\usepackage{nicefrac}
\usepackage{mathrsfs}
\usepackage{xcolor}
\usepackage{colortbl}
\usepackage{marvosym}
\usepackage[export]{adjustbox}
\usepackage{pifont}
\usepackage{cleveref}
\usepackage{ifpdf}
\ifpdf
  
\else
\fi
\usepackage{color}
\definecolor{cyan}{cmyk}{1,0,0,0}
\definecolor{darkgreen}{rgb}{0,0.5,0}
\definecolor{orange}{rgb}{1,0.5,0}
\definecolor{magenta}{cmyk}{0,1,0,0}
\definecolor{darkyellow}{cmyk}{0,0,0.75,0}
\definecolor{gray}{rgb}{0.8,0.8,0.8}
\hypersetup{colorlinks=true, linkcolor=cyan, filecolor=blue, urlcolor=red, citecolor=green,}
\usepackage{fancyhdr}
\renewcommand\footnotetextcopyrightpermission[1]{}
\copyrightyear{2023} 
\acmYear{2023} 
\setcopyright{acmlicensed}
\acmConference[MM '23]{Proceedings of the 31th ACM International Conference on Multimedia}{October 29–-November 3, 2023}{Ottawa, Canada}
\acmBooktitle{Proceedings of the 31th ACM International Conference on Multimedia (MM '22), October 29–-November 3, 2023, Ottawa, Canada}
\acmSubmissionID{906}

\begin{document}
\setcitestyle{numbers,sort&compress}

% \title{Perception-Distortion Balanced Image Denoising Based on Diffusion Models with Adaptive Noisy Image Embedding}
% \title{Perception-Distortion Balanced Image Denoising Based on Pre-trained Diffusion Models with Adaptive Embedding}
% \title{Image Denoising Based on Diffusion Models via Adaptive Embedding and Ensembling}

\title{Mutual-Guided Dynamic Network for Image Fusion}

\author{Yuanshen Guan}
\affiliation{%
  \institution{University of Science and Technology of China}
   \city{Hefei}
   \state{Anhui}
   \country{China}}
\email{guanys@mail.ustc.edu.cn}

\author{Ruikang Xu}
\affiliation{%
  \institution{University of Science and Technology of China}
   \city{Hefei}
   \state{Anhui}
   \country{China}}
\email{xurk@mail.ustc.edu.cn}

\author{Mingde Yao}
\affiliation{%
  \institution{University of Science and Technology of China}
   \city{Hefei}
   \state{Anhui}
   \country{China}}
\email{mdyao@mail.ustc.edu.cn}

\author{Lizhi Wang}
\affiliation{%
  \institution{Beijing Institute of Technology}
  \city{Beijing}
  \country{China}}
\email{wanglizhi@bit.edu.cn}

\author{Zhiwei Xiong$^{\dag}$}
\thanks{$^{\dag}$Corresponding author}
\affiliation{%
  \institution{University of Science and Technology of China}
   \city{Hefei}
   \state{Anhui}
   \country{China}}
\email{zwxiong@ustc.edu.cn}

\begin{abstract}

Image fusion aims to generate a high-quality image from multiple images captured under varying conditions. The key problem of this task is to preserve complementary information while filtering out irrelevant information for the fused result. However, existing methods address this problem by leveraging static convolutional neural networks (CNNs), suffering two inherent limitations during feature extraction, \textit{i.e.}, {being unable to handle spatial-variant contents and lacking guidance from multiple inputs}. In this paper, we propose a novel mutual-guided dynamic network (MGDN) for image fusion, which allows for effective information utilization across different locations and inputs. Specifically, we design a mutual-guided dynamic filter (MGDF) for adaptive feature extraction, composed of a mutual-guided cross-attention (MGCA) module and a dynamic filter predictor, where the former incorporates additional guidance from different inputs and the latter generates spatial-variant kernels for different locations. In addition, we introduce a parallel feature fusion (PFF) module to effectively fuse local and global information of the extracted features. To further reduce the redundancy among the extracted features while simultaneously preserving their shared structural information, we devise a novel loss function that combines the minimization of normalized mutual information (NMI) with an estimated gradient mask. Experimental results on five benchmark datasets demonstrate that our proposed method outperforms existing methods on four image fusion tasks. The code and model are publicly available at: https://github.com/Guanys-dar/MGDN.

\end{abstract}

%    <concept>
%        <concept_id>10010147.10010178.10010224</concept_id>
%        <concept_desc>Computing methodologies~Computer vision</concept_desc>
%        <concept_significance>500</concept_significance>
%        </concept>
%  </ccs2012>

% \ccsdesc[500]{Computing methodologies~Computer vision}

\keywords{Image Fusion, Dynamic Filter, Mutual Information.}

\maketitle

\vspace{-2mm}
\section{Introduction}
Image fusion is an important task in computer vision that fuses multiple images captured under different conditions to create a single integrated image with abundant information. The key step in image fusion is to preserve complementary information while filtering out irrelevant information for the fused result. For instance, multi-focus image fusion (MFF)~\cite{SFMD,DRPL} aims to gather essential information from images of different focus depths into a single clear image without blur, and high-dynamic-range (HDR) deghosting~\cite{Sen12,kalaetal,AHDR} aims to obtain an HDR image with clear texture details while suppressing oversaturated and blurry artifacts from several low-dynamic-range (LDR) images. 

Various image fusion algorithms have been proposed over time, mainly falling into two categories: traditional methods and deep learning-based methods. Traditional methods~\cite{Sen12,HU13,MFF_old1,MFF_old2,MEF_old1,MEF_old2} typically rely on mathematical transformations and manually designed fusion rules, which face challenges when dealing with diverse source images. Recently, deep learning-based methods~\cite{IFCNN,U2Fusion,AHDR, Xurk_Stereo_1} have seen a flourishing development due to their powerful capability for feature extraction.
These methods typically use static convolutional filters~\cite{Alexnet} that are trained only once and then applied uniformly to all images.
However, image fusion tasks need to handle diverse inputs, such as images with spatial-variant contents and drastic changes across scenes. 

\begin{figure}[t]
  \centering
  \includegraphics[width=\linewidth]{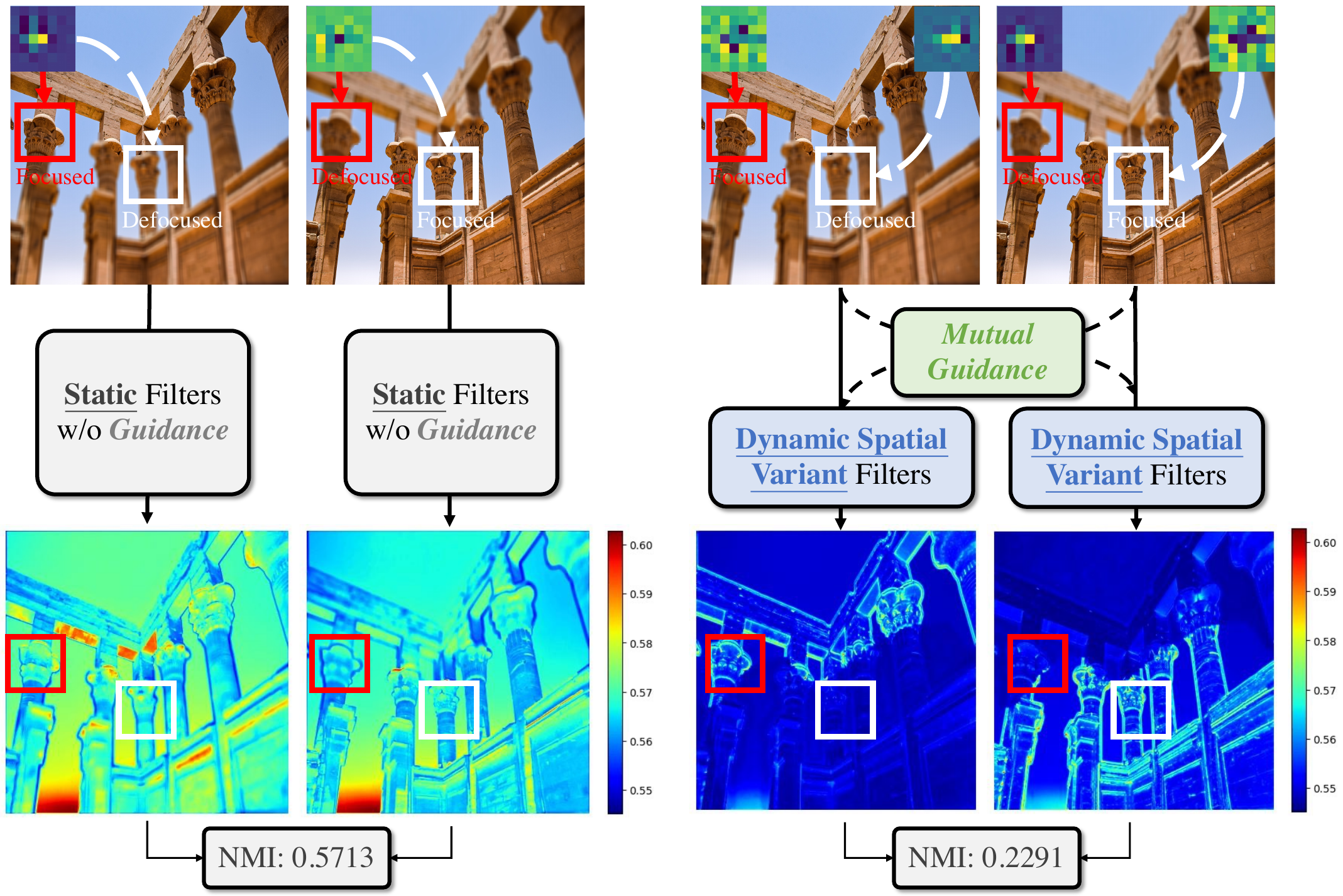}
  \vspace{-7mm}
  \caption{
  An example of multi-focus image fusion. Left: static filters treat all regions uniformly without consideration of the complementary contents contained in another input. Right: our dynamic filters and mutual guidance help the network remove blurry features while preserving clear content for final fusion. Our dynamic filtered features obtain a lower value of normalized mutual information (NMI)~\cite{NMI}, indicating that more complementary features are extracted.}
  \vspace{-5mm}
  \label{Fig:teaser}
\end{figure}

The existing static convolutional filter is inadequate in capturing spatial variance and scene changes, leading to limited adaptability and flexibility~\cite{BurstSR,jiaetal}. As shown in Fig.~\ref{Fig:teaser}, for the task of MFF, the static filter uniformly slides over the entire image, treating all pixels equally without taking into account the spatial variance of the depth-of-field at different locations, which leads to ineffective information utilization during feature extraction (e.g., the strong activation value in blurry areas). Additionally, the static filter merely pertains to a single input and lacks consideration of mutual guidance from multiple inputs, which is crucial to preserving complementary information and suppressing irrelevant information across different inputs. 

In this paper, we design a novel mutual-guided dynamic network (MGDN) for image fusion, which allows for effective information utilization across different locations and inputs. Specifically, we propose a mutual-guided dynamic filter (MGDF) for adaptive feature extraction, which consists of a mutual-guided cross-attention (MGCA) module and a dynamic filter predictor. The MGCA module enables the incorporation of additional guidance from different inputs, while the dynamic filter predictor generates spatial-variant kernels to accommodate different locations. The generated spatial-variant kernels are then applied for convolution, exhibiting remarkable adaptability in capturing spatial variance with guidance information. Moreover, we introduce a parallel feature fusion (PFF) module to effectively fuse local and global information of the extracted features. The whole network iteratively stacks MGDF and PFF with dynamic mutual guidance, achieving an effective and versatile architecture for image fusion.

To further mitigate the redundant information while preserving
the shared structure information among the extracted features, we devise a novel loss to minimize the normalized mutual information (NMI)~\cite{NMI} with an estimated gradient mask, which preserves more fine-grained structure information before complementary information extraction. Different from previous methods that apply the regression loss to final fused images, our devised loss directly constrains the intermediate features (\emph{i.e.,} the output of MGDF). This enables the filtering process to follow explicit regularity for discerning between complementary and common information.

We evaluate MGDN on four representative image fusion tasks, including multi-exposure image fusion (MEF), MFF, HDR deghosting, and RGB-guided depth map super-resolution (GDSR). The experimental results demonstrate that our method outperforms existing state-of-the-art (SOTA) image fusion methods on five benchmark datasets~\cite{MFFW,LYTRO,MEFB,kalaetal,NYUv2} in terms of various evaluation metrics. It is also worth noting that, benefiting from the mutual-guided dynamic filtering, the proposed MGDN not only performs well on well-aligned image fusion tasks but also has a generalization ability for misaligned inputs. In summary, our contributions are as follows:

\par {$\bullet$} We propose MGDN as a unified image fusion network, which effectively preserves complementary information and suppresses irrelevant information from multiple inputs. 
\par {$\bullet$} We propose MGDF for adaptive feature extraction as a core design of MGDN with two primary benefits: 1) dynamic filtering that adapts to different contents and 2) mutual guidance that effectively leverages information from different inputs. 
\par {$\bullet$} We devise a novel loss to minimize the mutual information with an estimated gradient mask on the feature level, which reduces the redundancy among extracted features while simultaneously preserving the shared structural information. 
\par {$\bullet$} Extensive experiments on five benchmark datasets demonstrate the superiority of our method compared to SOTA image fusion methods on four representative image fusion tasks.

\section{Related Work}
\subsection{Image Fusion}

As a pivotal technique in image processing, image fusion has gained increasing attention in recent years. Early works~\cite{oldfuse1,oldfuse2,SPDMEF,yao2019spectral,xu2021stereo} focus on designing task-specific fusion algorithms and achieve promising performance. Recently, researchers have made great progress in developing unified image fusion methods.
One notable contribution in this direction is MST-SR~\cite{MSTSR}, a pioneering image fusion method that integrates multi-scale transform and sparse representation techniques to aggregate complementary information. Subsequently, Zhang \emph{et al.}~\cite{IFCNN} propose CNN-based methods for image fusion based on DeepFuse~\cite{DeepFuse}, which remarkably improve the performance on various image fusion tasks. Another method, PMGI~\cite{PMGI}, considers different image fusion problems as proportional maintenance of gradient and intensity, and introduces a unified form of loss functions. Notably, recognizing that different fusion scenarios can mutually benefit each other, Xu \emph{et al.}~\cite{U2Fusion} devise a unified unsupervised image fusion model for multi-fusion tasks by combining learnable information measurement and elastic weight consolidation techniques~\cite{EWC1,EWC2}. In addition, Ma \emph{et al.}~\cite{swinfusion} introduce a joint mixed-architecture fusion network that exploits long-range interactions in images. However, existing fusion methods rely on static networks and do not account for spatial and scene variation, which is crucial for multiple image fusion tasks in real-world applications.

\begin{figure*}[t]
  \centering
  \includegraphics[width=0.98\linewidth,height=0.7\linewidth]{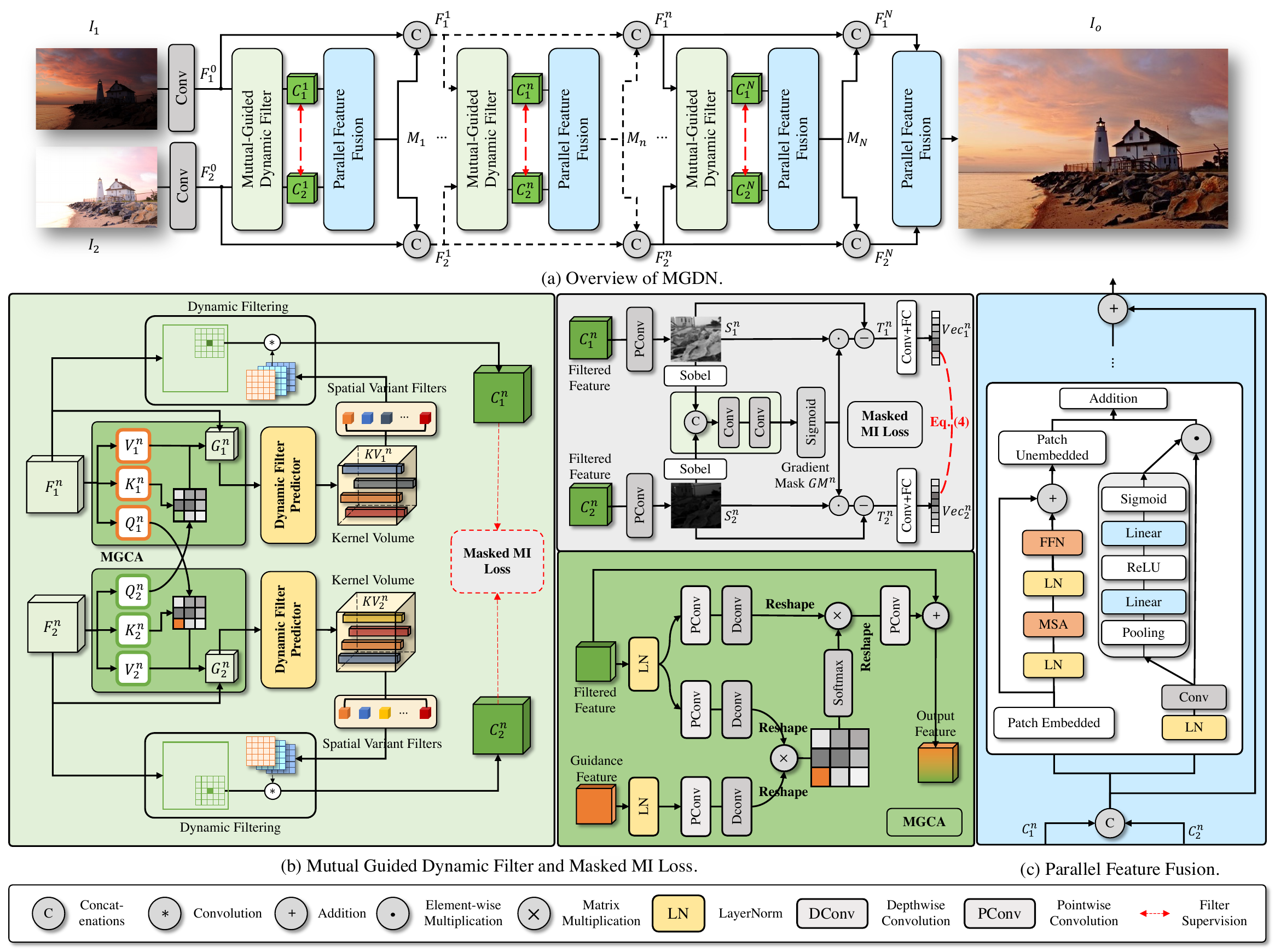}
  \vspace{-2mm}
  \caption{(a) Architecture of MGDN for image fusion. Our MGDN consists of a cascaded design that incorporates the MGDF and PFF modules. (b) MGDF consists of mutual-guided cross attention for aggregating information from different inputs and dynamic filter predictors to estimate the spatial-variant filters. Moreover, we propose masked MI loss to supervise the MGDF to suppress feature redundancy while preserving common structure. (c) PFF consists of dual-branch modules that incorporate channel attention blocks and shifted-window transformers to explore both global and local information for feature fusion.}
  \label{Fig:overview}
  \vspace{-5mm}
\end{figure*}

\vspace{-2mm}
\subsection{Dynamic Filtering}
Dynamic filtering, also referred to as filter adaptive processing, is a technique that involves learning per-pixel kernels, as opposed to using a single kernel across the entire image. The bilateral filter~\cite{BF1,BF2}, as a classic dynamic filtering method, processes a pixel via averaging its neighbors, weighted by the Gaussian of both spatial and intensity or color distances. He~\emph{et al.} proposed an method, called guided image filter~\cite{GIF}, which is a locally linear transform of the guidance image. Jia \emph{et al.}~\cite{jiaetal} first introduce the concept of dynamic filtering into the deep neural network, where filters are generated dynamically conditioned on an input and each location is filtered with a location-specific filter. After that, image adaptive filter has been successfully employed in various low-level computer vision tasks, such as video deblurring~\cite{deblurring}, super-resolution~\cite{DSR1,DSR2,Xurk_restoration,yao2022towards}, and defocus deblurring~\cite{defocus}. In addition, Transformer~\cite{swinir,swinfusion,YMD_Stereo_transformer} is a dynamic network that can be used for visual processing by calculating the relevance of different regions of the feature to establish long-range connections. In the context of image fusion tasks, processing multi-modal images necessitates a meticulously-designed filter module that can effectively suppress or eliminate undesired information while preserving the desired information. However, few attempts have been made in image fusion tasks to explore this kind of content-aware dynamic filtering.

\vspace{-1mm}
\section{Method}
In this section, we first depict the overview of the proposed image fusion method. Next, the design of the mutual-guided dynamic filter (MGDF) and the parallel feature fusion (PFF) are presented.

\vspace{-1mm}
\subsection{Overview}
Our network takes a set of images $ I_i \in \mathbb{R}^{H \times W \times C_{in}}, i \in \{1, 2, ...\}$ as inputs, and generates a fused image $I_o \in \mathbb{R}^{H \times W \times C_{out}}$ with informative scene representation, where $H, W$ are the height and width of the images, and $C_{in},C_{out}$ are channels numbers of input and output image.

As shown in Fig.~\ref{Fig:overview} (a), the proposed network consists of $N$ stages, where each stage contains an MGDF for filtering information and a PFF for fusing features. Without loss of generality, we take a pair of input images ($I_1, I_2$) as an example, and more inputs can be readily inferred by increasing $i$. To be detailed, the images ($I_1, I_2$) are first mapped to the feature domain via $3 \times 3$ convolutional layers, resulting in discriminative features ($F^0_1, F^0_2$). Then, MGDF extracts feature ($C_1^1, C^1_2$), which are dynamically obtained with mutual guidance from each other. Subsequently, the filtered features ($C_1^1, C^1_2$) are fed into PFF to merge global and local features in a parallel architecture, obtaining the merged feature $M^1$. The $M^1$ is then concatenated with the ($F_1^1, F^1_2$) and sent into the following MGDF and PFF for $N$ iterations. Finally, the $N^{th}$ PFF fuses the feature ($F^N_1, F^N_2$) and generates the fused image $I_o$.

\vspace{-4mm} 
\subsection{Mutual-Guided Dynamic Filter}

Existing static convolutional operations are incapable of spatially variable processing and are unaware of complementary content in other inputs. To address the aforementioned limitations, we propose MGDF, which consists of a mutual-guided cross attention (MGCA) and a dynamic kernel predictor to dynamically extract complementary information. We also propose a novel loss that constrains the output of MGDF to extract complementary information.

\begin{figure}[t]
  \centering
  \includegraphics[width=\linewidth]{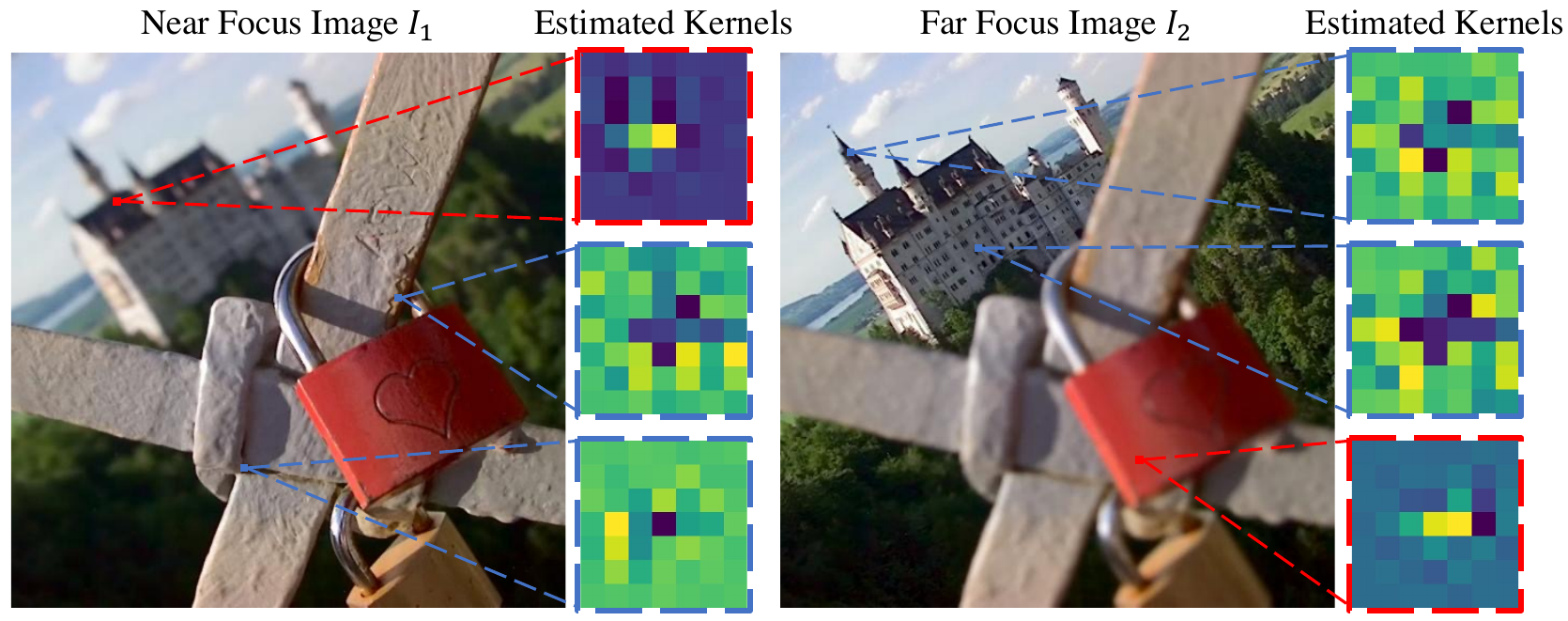}
  \vspace{-7mm}
  \caption{Illustration of dynamic filters in MFF. For near-focus image, the dynamic filter predictor estimates different kernel weights for focused and defocused areas. In the far-focus image, the predicted filter weights show the opposite pattern. The proposed dynamic filter estimates varying kernels for different image contents.}
  \label{filters_visualization}
  \vspace{-6mm}
\end{figure}

\subsubsection{Mutual-Guided Cross Attention} 
We propose MGCA to aggregate information from different inputs using a novel cross-attention mechanism, where the aggregated features are then utilized for dynamic filter prediction (see Sec.~\ref{DynamicFilter}). Different from the standard Self-Attention (SA) process~\cite{SA0,ViT,SwinT}, our proposed MGCA leverages the information from other inputs, enabling the extraction of complementary information with mutual guidance.

Specifically, for $n^{th}$ MGCA module, the input features $(F^n_1,F^n_2) \in \mathbb{R}^{H\times W\times C}$ are firstly fed to Layer Normalized (LN)~\cite{LN}. Then they are fed to point-wise convolutions to aggregate pixel-wise cross-channel context, followed by 3 $\times$ 3 depth-wise convolutions to encode channel-wise spatial context. The generated matrices are then reshaped to $\in \mathbb{R}^{HW\times C}$, yielding ($Q^n_1$, $K^n_1$, $V^n_1$) from $F^n_1$, ($Q^n_2$, $K^n_2$, $V^n_2$) from $F^n_2$. To allow guidance among features from different inputs, we exchange the generated query tokens among different inputs, resulting ($Q^n_2$, $K^n_1$, $V^n_1$) and ($Q^n_1$, $K^n_2$, $V^n_2$).
Without the loss of generality, we take the $F_1^n$ as the guidance feature as an example.
As shown in Fig.~\ref{Fig:overview}~(b), the exchanged query tokens $Q_1^n \in \mathbb{R}^{HW\times C}$ and the transposed raw key tokens $K_2^n \in \mathbb{R}^{C \times HW}$ calculate their dot-product interaction and generate a transposed attention map of size $\mathbb{R}^{C\times C}$, which refers to the similarity between the two distinct features, which is then used to select the relative feature in $V_2^n$. The selected feature are then reshaped back to $\in \mathbb{R}^{H \times W \times C}$ and added with the $F_2^n$. Moreover, the transposed calculation~\cite{restormer} averts the intensive calculation of standard SA. Overall, the MGCA process is defined as
% \vspace{-1mm}
\begin{equation}
\begin{split}
 & G^n_1=\operatorname{Attention}({Q}^n_2,{K}^n_1,{V}^n_1)+F^n_1, \\
 & G^n_2=\operatorname{Attention}({Q}^n_1,{K}^n_2,{V}^n_2)+F^n_2, \\
 & \operatorname{Attention}({Q}^n_1,{K}^n_2,{V}^n_2)={V}_2 \cdot \operatorname{Softmax}({K}^n_2 \cdot {Q}^n_1/\alpha), \\
 & \operatorname{Attention}({Q}^n_2,{K}^n_1,{V}^n_1)={V}_1 \cdot \operatorname{Softmax}({K}^n_1 \cdot {Q}^n_2/\alpha),
\end{split}
\end{equation}
where ($G^n_1$,$G^n_2$) is output of MGCA, $({Q}^n_1,{Q}^n_2) \in \mathbb{R}^{HW \times C}$, $({K}^n_1,{K}^n_2) \in \mathbb{R}^{C \times HW}$ and $({V}^n_1,{V}^n_2) \in \mathbb{R}^{HW \times C}$ are the reshaped tokens. Here, $\alpha$ is a learnable scaling parameter. Similar to the standard multi-head SA, we divide the number of channels into ``heads'' and learn separate attention maps parallelly. 

\subsubsection{Dynamic Filter Predictor}
\label{DynamicFilter}

To estimate spatial-variant dynamic kernels, we introduce a dynamic filter predictor to predict corresponding dynamic kernel volumes. Specifically, given the feature $(G^n_1,G^n_2) \in \mathbb{R}^{H \times W \times C} $ generated from MGCA, the dynamic filter predictor generates kernel volumes $(KV_1^n,KV_2^n) \in \mathbb{R}^{H \times W \times k^2}$, which represents convolution kernels with the size of $k\times k$ for each pixel. The kernel volumes are then reshaped and convolved with the feature ($F_1^n$, $F_2^n$) (\emph{i.e.,} the input of MGDF) in a spatial-variant way, obtaining the output feature $C_i^n$ as

\begin{figure}[t]
  \label{Kernel_Supervision}
  \centering
  \includegraphics[width=\linewidth]{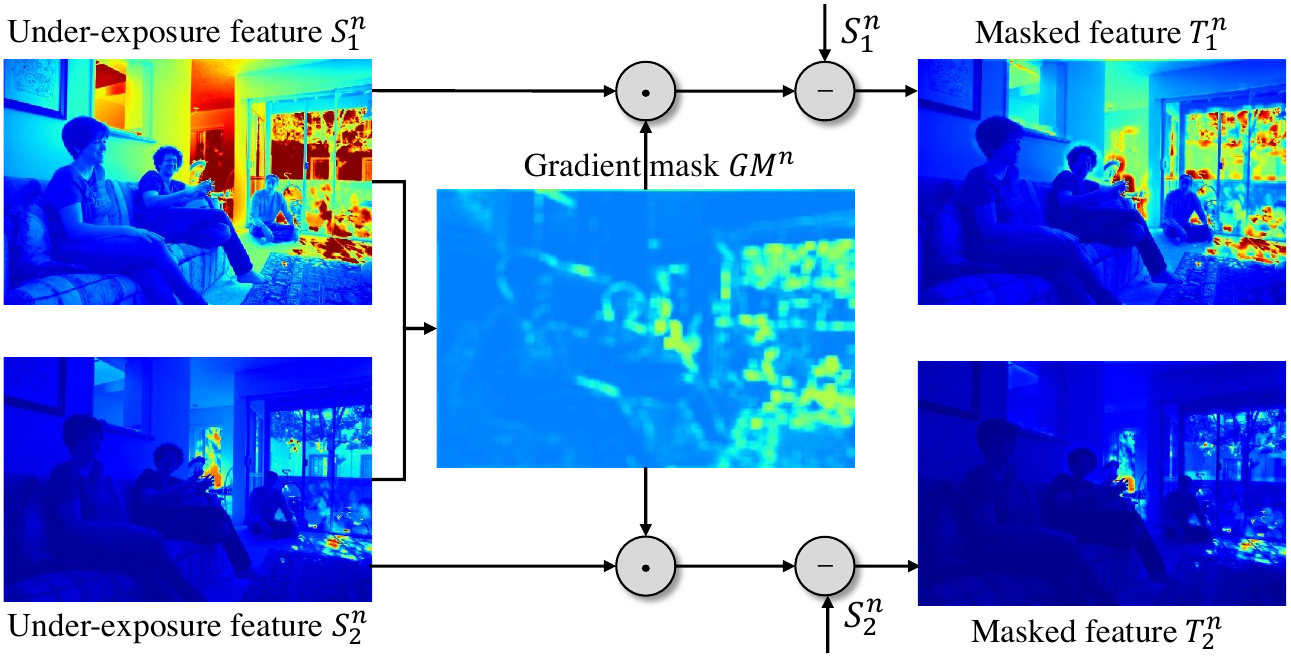}
  \vspace{-7mm}
  \caption{The estimated gradient mask in HDR deghosting for masked MI Loss calculation. The mask indicates the regions containing more common structures with a higher value. The mask protects the relative information, while the MI minimization reduces the redundancy among different inputs.}
  \label{Fig:fig4}
  \vspace{-5mm}
\end{figure}

\vspace{-1mm}
\begin{equation}
C_1^n = F_1^n \ast KV^n_1, \quad
C_2^n = F_2^n \ast KV^n_2, 
\end{equation}
where $\ast$ is the convolutional operator. Note that we share the kernel parameters along the channel dimension and perform the convolution in a depth-wise form~\cite{dwconv} to reduce the computational cost of dynamic filtering.

\begin{figure*}[tb]
  \centering
  \includegraphics[width=0.98\linewidth]{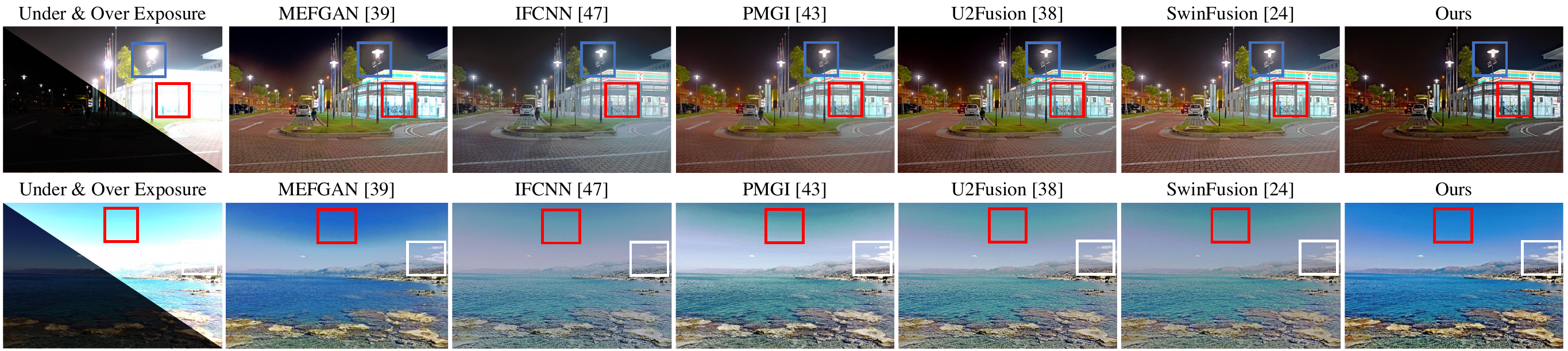}
  \vspace{-1mm}
  \caption{Visual comparison of MEF on benchmark dataset~\cite{MEFB}.}
  \vspace{-3mm}
  \label{Fig:MEFvisual}
\end{figure*}

\begin{table*}[tb]
  \caption{Quantitative comparisons of MEF on benchmark dataset~\cite{MEFB}. \textbf{\textcolor{red}{Red}} color indicates the best results and \textbf{\underline{\textcolor{blue}{Blue}}} color indicates the second-best results. ``$\downarrow$'' means that the smaller the index value, the better, and vice versa.}
  \vspace{-2mm}
  \label{Tab:MEF}
  \label{tab:performance}
  \resizebox{0.9\linewidth}{!}{
  \begin{tabular}{cccccccc}
    \toprule
    Methods & SPD-MEF~\cite{SPDMEF} & MEF-GAN~\cite{MEFGAN} & IFCNN~\cite{IFCNN} & PMGI~\cite{PMGI} & U2Fusion~\cite{U2Fusion} & SwinFusion~\cite{swinfusion} & MGDN (Ours) \\
    \midrule
    CE $\downarrow$ & 3.2146  & \underline{\textcolor{blue}{2.8222}} & 3.4098 & 3.0084 & 2.9761 & 2.8816 & \textcolor{red}{\textbf{2.3624}}\\
    SSIM $\uparrow$ & 0.9377  & 0.7722 & 0.9431 & 0.936 &  0.9304 & \textcolor{red}{\textbf{0.9538}} & \underline{\textcolor{blue}{0.9435}}\\
    $Q_{cb}$ $\uparrow$ & \underline{\textcolor{blue}{0.4551}}  & 0.3844 & 0.4112 & 0.4208 & 0.4174 & 0.4143 & \textcolor{red}{\textbf{0.4883}}\\
    VIF $\uparrow$ & 0.7772 & 0.581  & 0.7016 & 0.8077 & \underline{\textcolor{blue}{0.8358}} & 0.7417 & \textcolor{red}{\textbf{0.8538}}\\
    \bottomrule
  \end{tabular}}
  \vspace{-3mm}
\end{table*}

Different from vanilla dynamic filtering~\cite{jiaetal,defocus,DSR1}, our dynamic filter predictor leverages the information from other inputs, which adaptively incorporates the complementary information and suppresses the irrelevant information. We also visualize the predicted kernels in Fig.~\ref{filters_visualization}. It can be seen that, our MGDF can adaptively process each pixel and effectively extract complementary information, using variant kernels based on the content and guidance.

\vspace{-1mm}
\subsubsection{Masked MI Loss}
Building upon the proposed mutual-guided dynamic architecture, we delve deeper into a fundamental issue of image fusion: what types of information should be fused and suppressed to achieve optimal fusion results? Previous works~\cite{MI1,MIhigh} generally suppress feature redundancy and copy artifacts by minimizing mutual information (MI) loss for image fusion. However, directly applying MI loss can not only suppress the redundancy but also run the risk of discarding essential common characteristics. 

Consequently, as shown in Fig.~\ref{Fig:overview} (b), we propose masked MI loss to reduce feature redundancy while preserving fundamental relative structures in the filtered features. Specifically, given the filtered features ($C_1^n, C_2^n$), \emph{i.e.,} the outputs of MGDF, We first map them to one-channel features ($S_1^n, S_2^n$) via point-wise convolutions. Then we estimated a soft similarity mask $GM^n$ from their gradient features by several convolutions followed by a sigmoid function, where the gradient is extracted by classic Sobel filter~\cite{sobel}. Then the uncommon features ($T_1^n$, $T_2^n$) thus can be selected as

\vspace{-2em}
\begin{equation}
\begin{split}
% M^n = \operatorname{Sigmoid}(\operatorname{CNN}(\nabla{S_1^n}, \nabla{S_2^n}), \\
T_1^n = S_1^n \cdot (1-GM^n),\quad T_2^n = S_2^n \cdot (1-GM^n),
\end{split}
\end{equation}
where ($T_1^n$, $T_2^n$) are the selected uncommon one channel feature, as shown in Fig.~\ref{Fig:fig4}, which are then mapped to low-dimensional vectors ($Vec_1^n$,$Vec_2^n$) to calculate loss for complementary feature extraction. Subsequently, we introduce normalized mutual information (NMI) minimization to enforce the complementary information learning of two inputs. The masked MI loss $MaskMI(\cdot)$ is calculated as
\vspace{-1mm}
\begin{equation}
\begin{split}
& MaskMI(S_1^n,S_2^n) = 2* (1- \frac{H(Vec_1^n,Vec_2^n)}{H(Vec_1^n) + H(Vec_2^n)}),\\
& H(Vec_1^n) = -\sum{Vec_1^n\log(Vec_2^n)}-KL(Vec_1^n||Vec_2^n), \\
& H(Vec_2^n) = -\sum{Vec_2^n\log(Vec_1^n)}-KL(Vec_2^n||Vec_1^n), \\
\end{split}
\end{equation}
where $H(\cdot)$ represents the entropy, $H(Vec_1^n)$, $H(Vec_2^n)$ indicate the marginal entropies, and $H(Vec_1^n,Vec_2^n)$ is the joint entropy. Kullback-Leibler divergence $(\operatorname{KL(\cdot)})$ is used to calculate the entropy.

\vspace{-1mm}
\subsection{Parallel Feature Fusion}

We utilize a parallel feature fusion module to explore both global and local information for image fusion, as shown in Fig.~\ref{Fig:overview} (c). For the global branch, we employ the prowess of a window-based multi-head transformer encoder~\cite{swinir,HDRT}, adept at capturing long-range contexts. Meanwhile, the local branch employs a local context extractor to extract intricate local feature maps, enriched by channel attention~\cite{CA} to select the most valuable features.
The two branch features were combined with an element-wise addition. The output feature fuses global and local contexts, imbuing the fused result with meticulous details and comprehensive information.

\vspace{-1mm}
\subsection{Optimization}
The overall loss function consists of two parts: intensity L1 loss to maintain a proper intensity distribution and masked MI loss for extracting complementary information. For MEF~\cite{MEF_old1,MEF_old2}, MFF~\cite{MFF_old1,MFF_old2}, and GDSR~\cite{DSR1,DSR2}, the overall loss can be written as
\begin{equation}
L = ||I_o-GT||_1+\lambda\sum_{n=1}^N{MaskMI(S_1^n,S_2^n)},
\end{equation}
where $(S_1^n,S_2^n)$ denotes the one channel features mapped from the output of $n^{th}$ MGDF, $GT$ is the ground truth image and $\lambda$ is the parameters to balance the two terms in loss function. Specifically, for HDR deghosting, the network has three inputs. Thus, the masked MI loss is calculated twice: between filtered features of under-exposure LDR image and normal-exposure LDR image, and between filtered features of over-exposure image and normal-exposure LDR image. Loss for HDR deghosting $L_{HDR}$ can be written as
% Besides, following previous works~\cite{AHDR,HDRT}, the L1 loss is calculated on the tonemapped domain, and the tonemap function is denoted as $T(\cdot)$. 
\begin{equation}
\begin{split}
& L_{HDR} = ||I_o-GT||_1 + \lambda \sum_{n=1}^N{(MaskMI(S_2^n,S_1^n)+MaskMI(S_2^n,S_3^n))}.
\end{split}
\end{equation}

\vspace{-2.1em}
\section{Experiments}

% \subsection{Implementation Details}

% In this section, we compared our method with SOTA methods on four representative image fusion tasks. 
% % To evaluate the performance of MGDN, we carry out both quantitative and qualitative comparisons against these state-of-the-art methods. Furthermore, we conduct both quantitative and qualitative ablation studies to validate the effectiveness of the proposed design.
% We set the batch size to 4 and conduct 10,000 training steps for each fusion task. Training data are cropped randomly into 128 × 128 patches. Adam optimizer are used with an initial learning rate of $1\times10^{-4}$. More details can be found in our supplementary material.
% % All experiments are conducted on one NVIDIA GeForce RTX 3090Ti GPU. 

\begin{figure*}[t]
  \centering
  \includegraphics[width=0.98\linewidth]{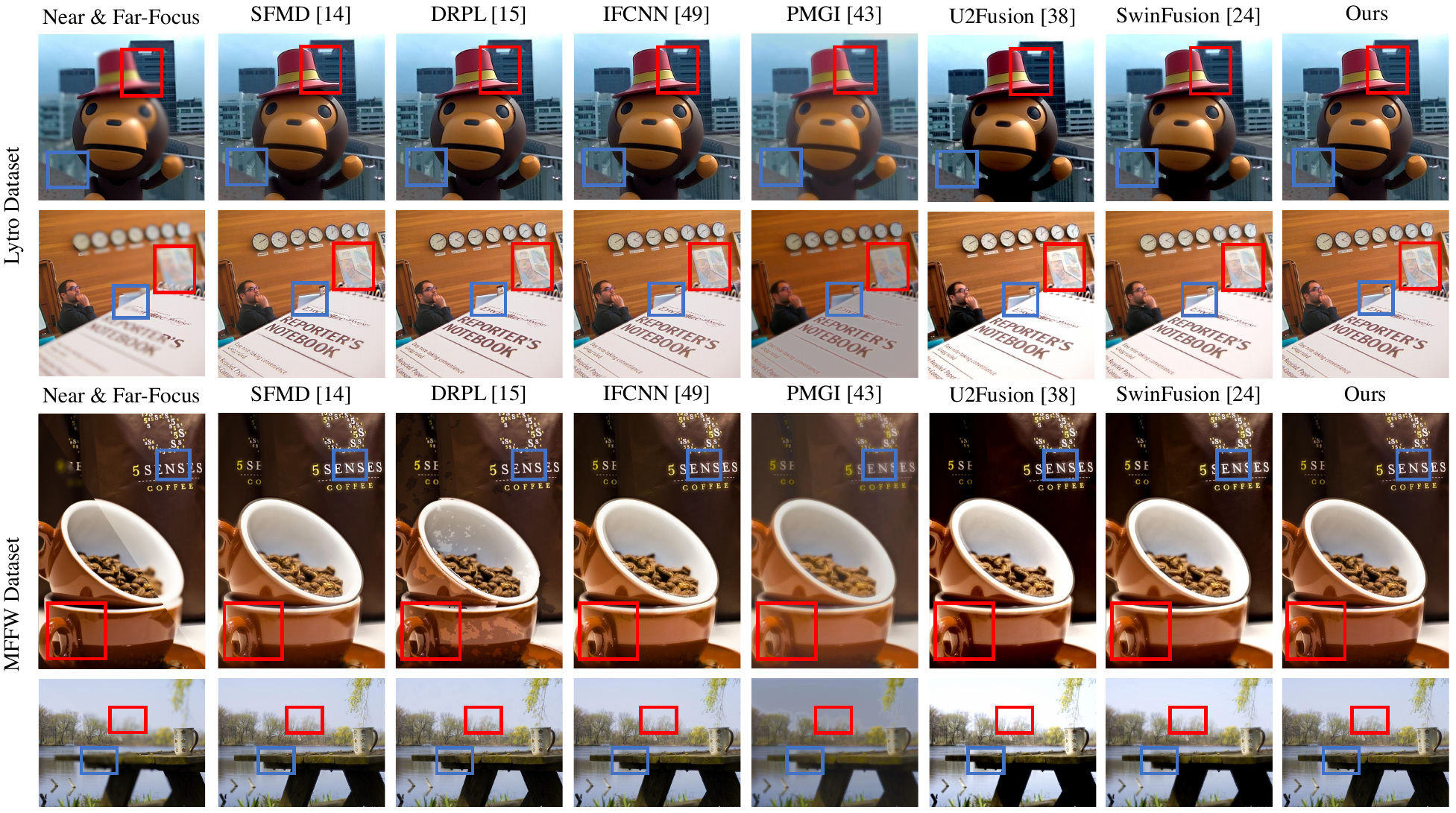}
  \vspace{-3mm}
  \caption{Visual comparison of MFF on Lytro~\cite{LYTRO} and MFFW dataset~\cite{MFFW}.}
  \label{Fig:MFFvisual}
  \vspace{-3mm}
\end{figure*}

\begin{table*}[t]
    \centering
    \caption{Quantitative comparisons of MFF on Lytro~\cite{LYTRO} and MFFW~\cite{MFFW} dataset. \textbf{\textcolor{red}{Red}} color indicates the best results and \textbf{\underline{\textcolor{blue}{Blue}}} color indicates the second-best results. Note that for all four metrics, higher values indicate better performance.}
    \vspace{-3mm}
    \resizebox{0.9\linewidth}{!}{
    \begin{tabular}{ccccccccccccc}
    \hline
        \toprule
        \multirow{2}{*}{Methods}& \multicolumn{4}{c}{Lytro~\cite{LYTRO}} & \multicolumn{4}{c}{MFFW~\cite{MFFW}} & \multicolumn{4}{c}{Aver} \\
         & $Q_{abf}$ & En & PSNR & SSIM & $Q_{abf}$ & En & PSNR & SSIM & $Q_{abf}$ & En & PSNR & SSIM \\ 
        \midrule
        SFMD~\cite{SFMD} & 0.609  & 7.526  & 64.542  & 0.841  & 0.554  & 7.110  & 62.365  & 0.798  & 0.587  & 7.362  & 63.658 & 0.824   \\ 
        DRPL~\cite{DRPL} & \textcolor{red}{\textbf{0.753}}  & 7.528  & 64.549 & 0.891  & \textcolor{red}{\textbf{0.710}} & \textcolor{red}{\textbf{7.143}}  & 62.734  & 0.781  & \textcolor{red}{\textbf{0.736}}  & \textcolor{red}{\textbf{7.376}}  & 63.834 & 0.848  \\ 
        IFCNN~\cite{IFCNN} & 0.725  & \textcolor{blue}{\underline{7.531}}  & \textcolor{blue}{\underline{64.645}}  & \textcolor{blue}{\underline{0.893}}  & 0.611  & 7.112  & \textcolor{blue}{\underline{63.160}} & \textcolor{blue}{\underline{0.816}} & 0.680  & 7.366  & \textcolor{blue}{\underline{64.060}} &\textcolor{blue}{\underline{0.863}}  \\ 
        PMGI~\cite{PMGI} & 0.386  & 7.273  & 62.292  & 0.858  & 0.367  & 6.948  & 60.549  & 0.746  & 0.378  & 7.145  & 61.065  & 0.814 \\ 
        U2Fusion~\cite{U2Fusion} & 0.609  & 7.345  & 62.234  & 0.841   & 0.539 & 6.835  & 61.110  & 0.768  & 0.581  & 7.144  & 61.791 & 0.812   \\ 
        SwinFusion~\cite{swinfusion} & 0.642  & 7.487  & 63.726  & 0.885  & 0.574  &  6.983  & 62.526  & 0.797  & 0.615  & 7.289  & 63.253 & 0.850 \\ 
        Ours & \textcolor{blue}{\underline{0.740}}  & \textcolor{red}{\textbf{7.533}}  & \textcolor{red}{\textbf{64.835}} & \textcolor{red}{\textbf{0.902}}  & \textcolor{blue}{\underline{0.701}}  & \textcolor{blue}{\underline{7.127}} & \textcolor{red}{\textbf{63.377}}  & \textcolor{red}{\textbf{0.839}}  & \textcolor{blue}{\underline{0.724}} & \textcolor{blue}{\underline{7.373}} & \textcolor{red}{\textbf{64.261}} & \textcolor{red}{\textbf{0.877}}   \\ 
        \bottomrule
    \end{tabular}}
    \vspace{-4mm}
    \label{Tab:MFF}
\end{table*}

\begin{figure*}[t]
  \centering
  \includegraphics[width=\linewidth]{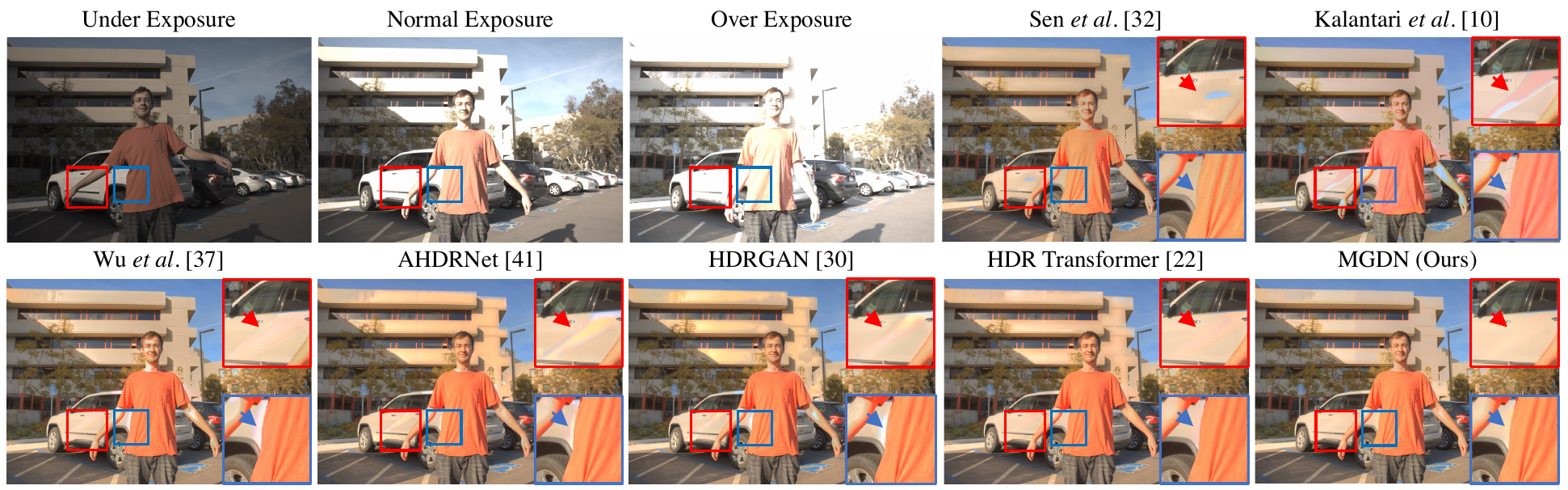}
  \vspace{-6mm}
  \caption{Visual comparisons for the HDR deghosting on the Kalantari~\emph{et al.}'s dataset~\cite{kalaetal}. (Readers are encouraged to refer to the electronic version of this document on a bright display for optimal visualization.)}
  \label{Fig:Deghosting}
  \vspace{-4mm}
\end{figure*}

\subsection{Multi-exposure Image Fusion}

\subsubsection{Experiment Configuration}
We use \cite{MEFTrain} as training set and~\cite{MEFB} as testset. We compare it with the following methods: SPD-MEF~\cite{SPDMEF}, MEFNet~\cite{MEFNet} and MEF-GAN~\cite{MEFGAN}, IFCNN~\cite{IFCNN}, PMGI~\cite{PMGI}, U2-Fusion~\cite{U2Fusion} and SwinFusion~\cite{swinfusion}. CE~\cite{CE}, SSIM~\cite{SSIM}, $Q_{cb}$~\cite{Qcb}, and VIF~\cite{VIF} are employed as test metrics.

\vspace{-1mm}
\subsubsection{Quantitative Comparisons}
The comparisons on the MEF are exhibited in Table~\ref{Tab:MEF}. From the results, we see that our method ranks first in the information theory-based metric CE~\cite{CE}, and the human visual perception metrics $Q_{cb}$~\cite{Qcb} and VIF~\cite{VIF}. The experimental results demonstrate that our proposed model successfully integrates complementary information and preserves texture information, resulting in visually pleasing fusion results.

\vspace{-1mm}
\subsubsection{Qualitative Comparisons}
The qualitative comparisons of MEF are shown in Fig.~\ref{Fig:MEFvisual}. As one can observe, other methods fail to maintain appropriate exposure levels since these methods lack guidance and spatial variance during the process of information processing. More specifically, the fusion results of SPD-MEF~\cite{SPDMEF} and MEFNet~\cite{MEFNet} lack overall and harmonious illumination and show obvious color distortions in the overexposed areas. The fusion results of MEFGAN~\cite{MEFGAN} and PMGI~\cite{PMGI} exhibit uneven brightness transitions. IFCNN~\cite{IFCNN} and U2Fusion~\cite{U2Fusion} have some issues with the overall color tone. Although the fusion results of SwinFusion~\cite{swinfusion} are relatively satisfactory, the fused information in the overexposed areas is not clear enough. Our MGDN can effectively merge the complementary information in the source images and maintain the appropriate exposure level through global exposure perception.

\vspace{-3mm}
\subsection{Multi-focus Image Fusion}

\subsubsection{Experiment Configuration}
The training set of MFF is composed of synthetic data used in DRPL~\cite{DRPL}, while the test set is the MFFW dataset~\cite{MFI-WHU} and Lytro~\cite{LYTRO} dataset. To verify the effectiveness of MFF, we compare it with the following methods: SFMD~\cite{SFMD}, DRPL~\cite{DRPL}, IFCNN~\cite{IFCNN}, PMGI~\cite{PMGI}, U2Fusion~\cite{U2Fusion} and SwinFusion~\cite{swinfusion}. $Q_{abf}$~\cite{Qabf}, Entorpy (En)~\cite{En}, PSNR~\cite{FusePSNR} and SSIM~\cite{SSIM} are employed as test metrics.

\vspace{-3mm}
\subsubsection{Quantitative Comparisons}
The quantitative comparisons between our method and other methods for MFF are exhibited in Table~\ref{Tab:MFF}. From the results, we see that our method ranks first in two information theory-based metrics, En and PSNR, and the structural similarity-based metric SSIM on the Lytro dataset~\cite{LYTRO}. Besides, the proposed method achieves the best PSNR and SSIM on the average result tested on both the Lytro dataset~\cite{LYTRO} and MFFW dataset~\cite{MFFW}. The above phenomena indicate that our method can effectively process and integrate information dynamically.

\begin{table}[tb]
\caption{Quantitative comparisons of HDR deghosting on Kalantari~\emph{et al.}'s dataset~\cite{kalaetal}. \textbf{\textcolor{red}{Red}} color indicates the best results and \textbf{\underline{\textcolor{blue}{Blue}}} color indicates the second-best results. For all four metrics, higher values indicate better performance}
\vspace{-3mm}
\label{Tab:Deghosting}
\begin{tabular}{lcccc}
\toprule
Methods & $\operatorname{PSNR}_{\mu}$ & $\operatorname{PSNR}_{L}$ & $\operatorname{SSIM}_{\mu}$ & $\operatorname{SSIM}_{L}$ \\
\midrule
Sen~\emph{et al.}~\cite{Sen12} & 40.97 & 38.34 & 0.9859 & 0.9764 \\
Kalantari~\emph{et al.}~\cite{kalaetal}& 42.72 & 41.22 & 0.9889 & 0.9829 \\
Wu~\emph{et al.}~\cite{Wuetal} & 41.99 & 41.66 & 0.9878 & 0.9860 \\
AHDRNet~\cite{AHDR} & 43.70 & 41.18 & 0.9905 & 0.9857 \\
HDRGAN~\cite{HDRGAN} & 43.85 & 41.65 & 0.9906 & 0.9870 \\
HDRTransformer~\cite{HDRT} & \textcolor{blue}{\underline{44.20}} & \textcolor{blue}{\underline{42.15}} & \textcolor{blue}{\underline{0.9918}} & \textcolor{blue}{\underline{0.9889}} \\
MGDN (Ours) & \textcolor{red}{\textbf{44.46}} & \textcolor{red}{\textbf{42.22}} & \textcolor{red}{\textbf{0.9919}} & \textcolor{red}{\textbf{0.9890}} \\
\bottomrule
\end{tabular}
\vspace{-6mm}
\end{table}

\vspace{-2mm}
\subsubsection{Qualitative Comparisons}
We also present the results of subjective comparison for MFF in Fig.~\ref{Fig:MFFvisual}. From the results, we can note that all methods could integrate information from the focused regions in different source images and generate an all-in-focus image. However, SFMD~\cite{SFMD}, PMGI~\cite{PMGI} and U2Fusion~\cite{U2Fusion} fail to retain a harmonious intensity distribution due to the lack of guidance between different inputs. SwinFusion~\cite{swinfusion}, despite having enhanced vision perception, is unable to recover certain precise details due to a lack of spatial-variant processing. Our method enables adaptive focus region awareness and maintains the proper intensity distribution through global context aggregation.

\vspace{-2mm}
\subsection{HDR Deghosting}
\subsubsection{Experiment Configuration}
For HDR deghosting, we train our network on the widely used Kalantari \emph{et al.}’s dataset~\cite{kalaetal}. Each sample from Kalantari \emph{et al.}’s dataset comprises three LDR images with exposure values of <-2, 0, +2>, or <-3, 0, +3>, as well as a ground truth HDR image. We compare the results of the proposed MGDN with several SOTA methods, which include Sen~\emph{et al.}~\cite{Sen12}, Kalantari~\emph{et al.}~\cite{kalaetal}, Wu~\emph{et al.}~\cite{Wuetal}, AHDRNet~\cite{AHDR}, HDR-GAN~\cite{HDRGAN}, and HDRTransformer~\cite{HDRT}. Specifically, with unaligned LDR inputs, our method doesn’t require any pre-alignment.

\begin{figure*}[t]
  \centering
  \includegraphics[width=0.98\linewidth]{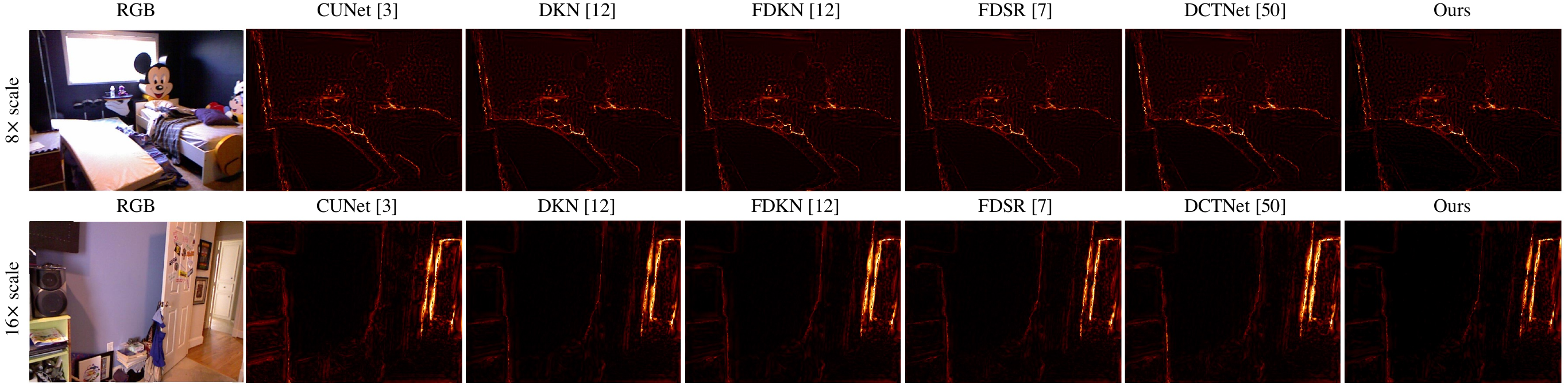}
  \vspace{-3mm}
  \caption{Visual comparison of error maps on GDSR on NYU v2 Dataset~\cite{NYUv2} for $\times$8, and $\times$16 SR scales.}
  \label{Fig:DepthSR}
  \vspace{-4mm}
\end{figure*}

\begin{table}
    \centering
    \caption{Quantitative comparison of GDSR between our MGDN and previous SOTA methods. We compare the RMSE metric (lower is better) on the NYU v2 dataset~\cite{NYUv2} for $\times$4, $\times$8, and $\times$16 SR scales.  \textbf{\textcolor{red}{Red}} color indicates the best results and \textbf{\underline{\textcolor{blue}{Blue}}} color indicates the second-best results.}
    \vspace{-3mm}
    \begin{tabular}{ccccc}
    \toprule
        Methods & NYU v2 $\times$4 & NYU v2 $\times$8 & NYU v2 $\times$16 \\
        \midrule
        DJFR~\cite{DJFR} & 2.38 & 4.94 & 9.18 \\ 
        PAC~\cite{PAC} & 1.89  & 3.33 & 6.78 \\ 
        CUNet~\cite{deng2020deep} & 1.92  & 3.70 & 6.78 \\
        DKN~\cite{DKN} & 1.62  & 3.26 & 6.51 \\ 
        FDKN~\cite{DKN} & 1.86  & 3.58 & 6.96 \\ 
        FDSR~\cite{FDSR} & 1.61  & 3.18 & 5.86 \\ 
        DCTNet~\cite{DCTNet} & \textcolor{blue}{\underline{1.59}}  & \textcolor{blue}{\underline{3.16}} & \textcolor{blue}{\underline{5.84}} \\
        MGDN (ours) & \textcolor{red}{\textbf{1.49}} & \textcolor{red}{\textbf{3.11}} & \textcolor{red}{\textbf{5.81}}\\
        \bottomrule
    \end{tabular}
    \vspace{-6mm} 
    \label{Tab:DepthSR}
\end{table} 

% \vspace{-1em}
\vspace{-1mm}
\subsubsection{Quantitative Comparisons}
We use PSNR and SSIM as evaluation metrics. To be more precise, we calculate $\operatorname{PSNR}_l$, $\operatorname{PSNR}_{\mu}$, $\operatorname{SSIM}_l$, and $\operatorname{SSIM}_{\mu}$ scores between the reconstructed HDR images and their corresponding ground truth. The '$L$' and '$\mu$' denote the linear and tonemapped domain values, respectively. Table~\ref{Tab:Deghosting} lists the quantitative results. The proposed MGDN surpasses the recently published HDR-Transformer~\cite{HDRT} by up to 0.26 dB and 0.07 dB in terms of $\operatorname{PSNR}_L$ and $\operatorname{PSNR}_{\mu}$, respectively, demonstrating the effectiveness of our method.

% \vspace{-1em}
\vspace{-2mm}
\subsubsection{Qualitative Comparisons}
All qualitative results are tone-mapped using the same Photomatix Pro parameters to ensure fair comparisons. Fig.~\ref{Fig:Deghosting} depicts an intractable scene with saturations and significant motion, with red and blue highlighting the two comparison locations, respectively. As can be seen, the red boxed area shows a large motion region caused by the hand. Previous methods remove some ghosting artifacts, but still produce colour distortions and inconsistent details. The blue boxed regions exhibit significant intensity variation, causing long-range saturation. Our method produces the highest-quality visual perception results with visually appealing scene colours.

% \vspace{-1em}
\vspace{-2mm}
\subsection{RGB-guided Depth Map SR}

\subsubsection{Experiment Configuration}
In this section, we assess the performance of MGDN on the NYU v2 dataset~\cite{NYUv2} and compare its results with several SOTA methods, including DJF~\cite{DJF}, DJFR~\cite{DJFR}, PAC~\cite{PAC}, CUNet~\cite{deng2020deep}, DKN~\cite{DKN}, FDKN~\cite{DKN}, FDSR~\cite{FDSR}, and DCTNet~\cite{DCTNet}. We evaluate the performance of GDSR against ground-truth depth maps in terms of root-mean-square error (RMSE). A lower RMSE indicates more accurate predicted depth maps. Note that the low-resolution depth map is upsampled by bicubic interpolation before being fed into the network.

\vspace{-1mm}
\subsubsection{Quantitative Comparisons}
The quantitative evaluation of our proposed MGDN is presented on the NYU v2 test sets, with SR factors of $\times$4, $\times$8, and $\times$16, as depicted in Table~\ref{Tab:DepthSR}. Our MGDN exhibits superior performance across various SR scales. This highlights the advantage of our model over SOTA methods.

\vspace{-1mm}
\subsubsection{Qualitative Comparisons}
Following previous works~\cite{DCTNet,DKN,FDSR}, we present a comprehensive analysis of error maps to compare depth images derived from different methods, as depicted in Fig.~\ref{Fig:DepthSR}. Qualitatively, our results demonstrate that the depth predictions generated by the MGDN have substantially fewer prediction errors and are more congruent with the ground truth images, highlighting the superior performance of the MGDN in GDSR.

\vspace{-1mm}
\subsection{Ablation Study}
We conduct ablation experiments about MGCA, dynamic filter predictor and masked MI Loss. In addition, ablation experiments about the PFF module can be found in the supplementary materials.

\begin{table}
    \centering
    \caption{Ablation study of the proposed MGDN. We compare the quantitive results on HDR deghosting.}
    \vspace{-3mm}
    \begin{tabular}{ccccc}
    \toprule
        Methods & $\operatorname{PSNR}_{\mu}$ & $\operatorname{PSNR}_L$ & $\operatorname{SSIM}_{\mu}$ & $\operatorname{SSIM}_L$ \\
        \midrule
        w/o MGCA & 44.16 & 41.75 & 0.9882 & \textcolor{blue}{\underline{0.9846}}\\ 
        w/o Dynamic Filter & 44.08 & 41.63 & 0.9840 & 0.9746\\
        w/o Masked MI Loss & \textcolor{blue}{\underline{44.23}} & \textcolor{blue}{\underline{41.77}} & \textcolor{blue}{\underline{0.9883}} & 0.9833\\
        % w/o Global Branch & 43.79 & 41.11 & 0.9893 & \textcolor{blue}{\underline{0.9854}}\\
        % w/o Local Branch & 44.15 & 41.67 & \textcolor{blue}{\underline{0.9899}} & 0.9846\\
        MGDN & \textcolor{red}{\textbf{44.46}} & \textcolor{red}{\textbf{42.22}} & \textcolor{red}{\textbf{0.9919}} & \textcolor{red}{\textbf{0.9890}} \\
        \bottomrule
    \end{tabular}
    \vspace{-6mm}
    \label{Tab:Ablation}
\end{table}

\begin{figure}[t]
  \centering
  \includegraphics[width=\linewidth]{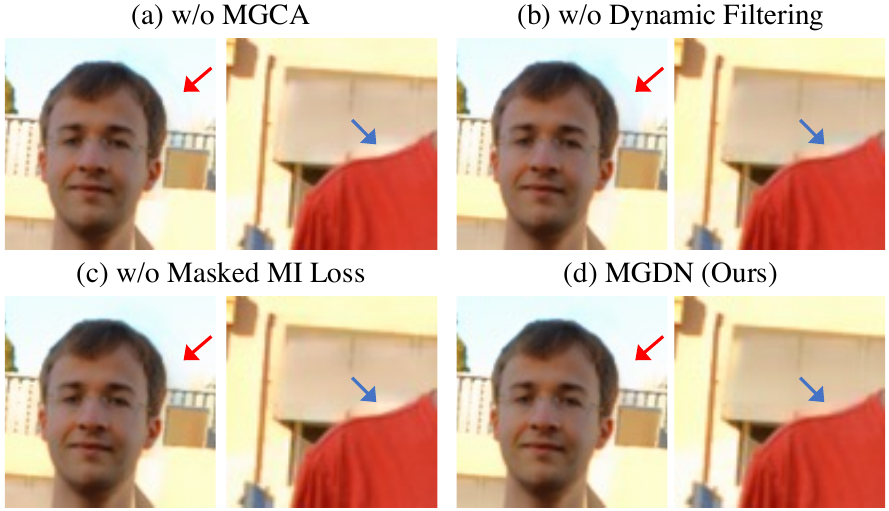}
  \vspace{-2em}
  \caption{Ablation experiments on HDR deghosting about the proposed designs.}
  \label{Fig:Ablation}
  \vspace{-5mm}
\end{figure}

% \subsubsection{Analysis on Mutual-Guided Cross Attention.}
\textit{Mutual-Guided Cross Attention.} For comparision, we replace MGCA with common transposed SA in the ablation experiment and train the network on the HDR deghosting task. As can be seen from Table~\ref{Tab:Ablation}, the performance drops 0.30 dB on $\operatorname{PSNR}_{\mu}$ without MGCA, depicting that the estimated kernel volume is blindness without the guidance. As shown in Fig.~\ref{Fig:Ablation}, we can witness areas with uneven color.

% \subsubsection{Analysis on Dynamic Filter Predictor.}
\textit{Dynamic Filter Predictor.} For comparision, we forced the convolution kernel weights of different spatial positions to be consistent while ensuring the same number of network parameters during the ablation experiment. As can be seen from Table~\ref{Tab:Ablation}, without dynamic filtering, the network performance drops severely, with 0.59 dB in $\operatorname{PSNR}_L$. As shown in Fig.~\ref{Fig:Ablation}, 
% the absence of guidance information can result in irregular local coloration.
static convolution, which disregards local information, performs inadequately in regions with motion.

% \subsubsection{Analysis on Mask MI Loss.}
\textit{Masked MI Loss.} For comparision, MGDN was trained solely under the supervision of L1 loss as a comparison. As can be seen from Table~\ref{Tab:Ablation}, the absence of masked MI loss leads to a performance drop, with 0.45 dB in $\operatorname{PSNR}_L$. As shown in Fig.~\ref{Fig:Ablation}, perceptible artifacts can be observed in challenging deghosting tasks due to the ineffective removal of irrelevant information during feature extraction.

% \vspace{-2mm}
\section{Conclusion}
In this paper, we introduce MGDN, a novel network for image fusion that effectively utilizes information from spatial-variant contents and inputs. Our proposed method leverages the MGDF module, which adaptively processes image features by taking into account both the current content and guided inputs. This module is composed of the MGCA for the integration of additional guidance information from different inputs, and the dynamic filter predictor for the estimation of spatial-variant filters for different locations, respectively. In addition, we introduce a PFF module to effectively combine local and global information from the extracted features. To address the issue of feature redundancy while preserving relative structural information, we propose a masked MI loss to supervise the filtering process. Our proposed method achieves SOTA performance on four image fusion tasks and five benchmark datasets, demonstrating its effectiveness in fusing images. Furthermore, we believe this method can be extended and applied to a wide range of image processing tasks.

% \section*{Acknowledgments}
% This work was supported in part by the National Natural
% Science Foundation of China under Grants 62131003 and
% 62021001.

\bibliographystyle{ACM-Reference-Format}
\bibliography{acmart}
\end{document}